\documentclass[a4]{article}

\usepackage{graphicx}
\usepackage{amsmath} 
\usepackage{enumitem}
\usepackage{caption}

\newcommand{\keywords}[1]{\textbf{\textit{Keywords---}} #1}

\begin{document}
\title{Comparison of Modified Kneser-Ney and Witten-Bell Smoothing
Techniques \\in Statistical Language Model \\of Bahasa Indonesia}
\author{Ismail \\
Computer Engineering Department \\
School of Applied Science, Telkom University \\
Bandung, Indonesia \\
ismailrusli@telkomuniversity.ac.id}

\date{}
\maketitle

\begin{abstract}
    Smoothing is one technique to overcome data sparsity
    in statistical language model. Although in its mathematical
    definition there is no explicit dependency upon
    specific natural language, different natures of 
    natural languages result in different effects
    of smoothing techniques. This is true for
    Russian language as shown by Whittaker\cite{whittaker1998comparison}.
    In this paper, We compared 
    Modified Kneser-Ney and Witten-Bell
    smoothing techniques in statistical 
    language model of Bahasa Indonesia.
    We used train sets of totally 22M words
    that we extracted from Indonesian version of Wikipedia. As far
    as we know, this is the largest train set used to
    build statistical language model for Bahasa Indonesia.
    The experiments with 3-gram, 5-gram, and 7-gram
    showed that Modified Kneser-Ney consistently outperforms
    Witten-Bell smoothing technique in term of 
    perplexity values. 
    It is interesting to note
    that our experiments showed
    5-gram model for Modified Kneser-Ney smoothing 
    technique outperforms that of 7-gram. Meanwhile, Witten-Bell
    smoothing is consistently improving over the increase of
    $n$-gram order. 

\end{abstract}

\section{Introduction}
Statistical language model is state-of-the-art
in modeling natural language. The model is used in application
such as automatic speech recognition, information
retrieval, and machine translation.
A statistical language model tries to estimate 
probability of a sentence to appear in a
modeled natural language.

To build statistical language model, 
first we have to prepare a corpus.
In English language, there are several standard
corpora, e.g. Brown corpus, WSJ (Wall Street Journal) corpus,
and NAB (North American Business news) corpus.
With these corpora, researchers in language model could benchmark
their works.

In Bahasa Indonesia, we hardly found a standard
corpus which is also freely available.
Therefore, we used here body of text available for free from
Wikipedia\footnote{dumps.wikimedia.org/idwiki/}.
We extracted the text from Wikipedia dump file using scripts available
in the Internet\footnote{blog.afterthedeadline.com/2009/12/04/generating-a-plain-text-corpus-from-wikipedia/}.
Before we used it, we preprocessed the text
to clean the noise as much as we can.
The resulting text has approximately 36M words
in 1,7M sentences. 

From the text, we build language models that
are smoothed using Modified Kneser-Ney and Witten-Bell 
techniques. Those techniques are among
the most widely used techniques in
languge model\cite{chen1996empirical}.
We used tools from SRI International\footnote{www.sri.com}, i.e. 
SRI Language Model
(SRILM)\footnote{www.speech.sri.com/projects/srilm/} to build our
language models. The tools 
are available
for free for non-commercial projects.

Our motivation in this work is based on the fact that we
did not find any studies on various smoothing techniques
in language model for Bahasa Indonesia.
Smoothing technique could improve
performance of any system in natural
language processing that uses statistical
language model. 
Although in its mathematical definition
of statistical language model there is no
dependency on specific natural language, the nature
of the language itself is different across languages.
For example, Russian language has large inflectional
words which gives rise to different results 
in smoothing techniques compared to that
of English language\cite{whittaker1998comparison}.
Therefore, our contribution here is to present
the study of two smoothing techniques, i.e., 
Modified Kneser-Ney and Witten-Bell, applied to
statistical language model of Bahasa Indonesia.
Various other smoothing techniques will be
studied in our future research.

This paper continues as follows. In section 2,
we mentioned several works in natural language processing
that used statistical language model of Bahasa Indonesia.
We did not find 
any explicit reference to smoothing techniques employed
in those works. 
We also pointed to the sizes of corpus used in the works
which are relatively
small compared to that we used here.
We proceed in section 3 describing statistical language
model and their smoothing techniques, i.e.
Modified Kneser-Ney and Witten-Bell techniques.
Section 4 detailed our experiments and results
which then concluded in section 5.

\section{Previous Works}
For English language model, Chen \textit{et al.}\cite{chen1996empirical}, 
had reported excellent study on various smoothing techniques.
They also proposed a modified version of
known Kneser-Ney smoothing techniques which is
slightly better from the original one. 

Teh\cite{teh2006hierarchical} used hierarchical Bayesian
model to build a language model and reported that
its performance is comparable to modified
Kneser-Ney technique.

As far as we know, there are no previous works
study various smoothing techniques for
Indonesian language model. 
We also noted that researches in natural language processing
which used statistical Indonesian language model 
do not refer to any smoothing techniques in their works.

Furthermore, we noted that there are no standard corpora 
to benchmark various language models for Bahasa Indonesia
that are also available for free.

One of the attempt to build corpus for Bahasa Indonesia
was reported by Larasati\cite{larasati2012identic}.
It is morphologically enriched Indonesia-English
parallel corpus with approximately 1M words for each language.

Sakti\cite{sakti2008development}, in buildng Large
Vocabulary Continues Speech Recognition (LVCSR) System,
had used different corpora, i.e., Daily News Task
(600K sentences), Telephone Application Task (2.5K sentences),
and Basic Travel Expression Corpus or BTEC Task (160K sentences).

Pisceldo \cite{pisceldo2009probabilistic} used 14K words size corpus
to build probabilistic
part-of-speech tagging in Bahasa Indonesia.

Riza \cite{riza2008ets} building Indonesian-English Machine Translation System
using Collaborative P2P corpus. This corpus has 280K sentences
with target 1 million sentence-pair (multilingual corpus)

For all corpora used in previous mentioned
works, none is available for free. The only corpus we found
freely available in the Internet is in Pan Localisation
Site\footnote{panl10n.net/english/OutputsIndonesia2.htm\#Linguistic\_Resources\_}
which is a work by Adriani\cite{adrianiresearch}. This corpus
consists of 500K words.

As mentioned earlier, we used body of text with size
of 36M words in our research. This is larger than that of corpora
used in previous mentioned works (although we admitted here
that our body of text is not tuned into a specific topic).
We do not claim that body of text we used here is standard.
It is our future goal to make it standard and
freely available in the future.

\section{Language Model}
A language model gives an answer to question
``what is the probability of sentence $W$ appears in a natural
language?''. Sentence like ``Saya suka buah'' 
(or in English, ``I like fruits'') is
naturally more probable to appear in a text
or everyday conversation than grammatically similar
sentence ``Saya suka meja'' (``I like table''). 

Formally, a language model approximates the probabilites
of a sequence of words $W = w_1w_2w_3\cdots w_i = w_{1}^{i}$.

\begin{equation}
    P(W) = P(w_{1}^{i})
    \label{eq:lm}
\end{equation}

Equation (\ref{eq:lm}) equals to find probability of 
the last word of the sentence given the previous words 
in the sentence times the probability of  
previous words in the sentence.

\begin{equation}
    P(W) = P(w_{1}^{i-1}) \times P(w_i|w_{1}^{i-1})
    \label{eq:lm_conditional}
\end{equation}

By chain rule we have

\begin{equation}
    P(W) = \prod_{j=1}^{i} P(w_{j}|w_{1}^{i-1})
    \label{eq:lm_broken}
\end{equation}

Using Markov property, we can assume that
probability of a word depends only on previous
$n-1$ words. This is $n$-gram language model
and we get

\begin{equation}
    P_{n}(W) = \prod_{j=1}^{i} P(w_{j}|w_{j-n+1}^{j-1})
    \label{eq:lm_ngram}
\end{equation}
when it is understood that in $w_k^l$, if 
$l<k$ the word is discarded.

To calculate the conditional probability in (\ref{eq:lm_ngram}),
we use maximum likelihood estimation. That is, 

\begin{equation}
    P_n(w_{i}|w_{i-n+1}^{i-1}) =
    \frac{\text{c}_n(w_{i-n+1}^{i})}{\text{c}_n(w_{i-n+1}^{i-1})}
    \label{eq:max_likelihood}
\end{equation}

with $\text{c}_n(w^i_{i-n+1})$ is the count of 
$w^i_{i-n+1}$ in training set.

The most common metric for evaluation 
a language model is perplexity\cite{chen1996empirical}.
Perplexity is defined by $2^H$, where $H$ is 
cross-entropy of the test set.

\begin{equation}
    H(T) = -\frac{1}{W_T}\text{log}_{2}P(T)
    \label{eq:cross-entropy}
\end{equation}

where $W_T$ is number of words in test set $T$.

A model is relatively better when it has lower
perplexity compared to that of other models.

\subsection{Smoothing Techniques}
Previous model suffers from one problem. It gives zero probability
to sentences that are not appear in the corpus. To overcome this, we
discount probability from sentences appear in the corpus
and distribute it to sentences that have zero probability. This 
techniques is called smoothing.

There are several smoothing techniques, e.g., Katz smoothing,
Good-Turing, Witten-Bell, and Kneser-Ney. 
Generally, those smoothing techniques fall into two
categories, backoff and interpolated techniques. 

In backoff technique, probability of sentences that are not appear
in corpus are estimated using that of its lower $n$-grams. 

\begin{equation}
    \begin{split}
        P_n^{BO} (w_i|w_{i-n+1}^{i-1}) = 
    \begin{cases}
            \tau_n(w_i|w_{i-n+1}^{i-1}) \\
            \qquad \text{if c}_n(w_{i-n+1}^i) > 0 \\ \\
        \gamma_n(w_{i-n+1}^{i-1})P_{n-1}^{BO}(w_i|w_{i-n+2}^{i-1}) \\ 
            \qquad \text{if c}_n(w_{i-n+1}^i) = 0 
    \end{cases}
    \end{split}
    \label{eq:backoff}
\end{equation}
where $\tau_n$ is modified probability for sentences appear
in corpus and 
$\gamma_n$ is scaling factor chosen to make
the backoff probability of sentences that do not appear
in corpus sum to one.

Instead of using backoff probability,
interpolated technique combines probability of a sentence
with that of its lower order, e.g. combined probability of
trigram, bigram, and unigram for trigram language model.

\begin{equation}
    \begin{split}
        P_n^I (w_i|w_{i-n+1}^{i-1}) = 
        \tau(w_i|w_{i-n+1}^{i-1}) + \\ 
        \gamma(w_{i-n+1}^{i-1}) P_{n-1}^I(w_i|w_{i-n+2}^{i-1}) 
    \end{split}
    \label{eq:interpolated}
\end{equation}

In this paper, we used Modified Kneser-Ney and Witten-Bell
and used backoff version of those techniques.

\subsection{Modified Kneser-Ney}
Initially, Kneser-Ney smoothing uses backoff technique. Chen and Goodman
modify it to use interpolation technique and further modify it
to have multiple discounts. This is called modified Kneser-Ney smoothing
technique\cite{chen1996empirical}.

\begin{equation}
    \tau_n(w_i|w_{i-n+1}^{i-1}) = \frac{c(w_{i-n+1}^i) - D(c(w_{i-n+1}^i))}  
        {\sum_{w_i}c(w_{i-n+1}^i)}
    \label{}
\end{equation}

\begin{equation}
    \begin{split}
    \gamma_n(w_{i-n+1}^{i-1}) =
    \frac{\sum_{j=1}^3 D_jN_j(w_{i-n+1}^{i-1}\text{\textbullet})}  
        {\sum_{w_i}c(w_{i-n+1}^i)}
    \end{split}
    \label{}
\end{equation}

$D$ is discounting values which is applied to sentences with
nonzero probabilities.
$N_j(w_{i-n+1}^{i-1}\text{\textbullet}) = |\{w_i|c(w_{i-n+1}^{i-1}) = j\}|$
is a number of words that appear after the context $w_{i-n+1}^{i-1}$
exactly $j$ times.
Modified Kneser-Ney used 3 different
discounting values $D_1$, $D_2$, and $D_{3+}$ which are discounting
value for $n$-grams with one, two, and three of more counts,
respectively.

\subsection{Witten-Bell}
To get the $\gamma$ value, Witten-Bell technique 
considers the number of unique words following
the history $w^{i-1}_{i-n+1}$. This number is formally
defined as

\begin{equation}
    N_{1+}(w_{i-n+1}^{i-1}\text{\textbullet}) =
        |\{w_i:c(w_{i-n+1}^{i-1}w_i) > 0\}|
    \label{eq:nplus}
\end{equation}

With this number, $\gamma$ parameter is defined as

\begin{equation}
    \gamma_n(w_{i-n+1}^{i-1}) =
    \frac{N_{1+}(w_{i-n+1}^{i-1}\text{\textbullet})}
        {N_{1+}(w_{i-n+1}^{i-1}\text{\textbullet}) +
            \sum_{w_i}c(w_{i-n+1}^i)}
    \label{eq:lwittenbell}
\end{equation}

and higher order distribution is defined as follow.

\begin{equation}
    \tau_n(w_i|w_{i-n+1}^{i-1}) = \frac{c(w_{i-n+1}^i)} 
        {N_{1+}(w_{i-n+1}^{i-1}\text{\textbullet}) +
            \sum_{w_i}c(w_{i-n+1}^i)}
    \label{eq:hwittenbell}
\end{equation}

\section{Experiment and Results}
We extracted body of texts from Indonesian 
version of Wikipedia. 
Before we fed those texts into SRILM, we preprocessed
them with following steps. 
\begin{enumerate}[parsep=0pt,itemsep=0pt]
    \item Removing unwanted lines
    \item Removing punctuations
    \item Splitting paragraphs into sentences
    \item Shuffle and remove duplicate sentences
    \item Remove sentences contain less than two tokens
\end{enumerate}
Table \ref{tab:corpus_stat} shows the number of words and sentences in resulting texts.

\begin{table}
    \centering
    \caption{Number of words and sentences in text}
    \begin{tabular}{| l | c | c | c | c |} \hline 
        Text & Indonesian \\ \hline \hline
       Sentences & 1,779,525 \\ \hline 
       Words & 36,597,737 \\ \hline
    \end{tabular}
    \label{tab:corpus_stat}
\end{table}

After we preprocessed the text, we did following steps
to prepare the data. 

\begin{figure}
 \centering
 \includegraphics[width=10cm]{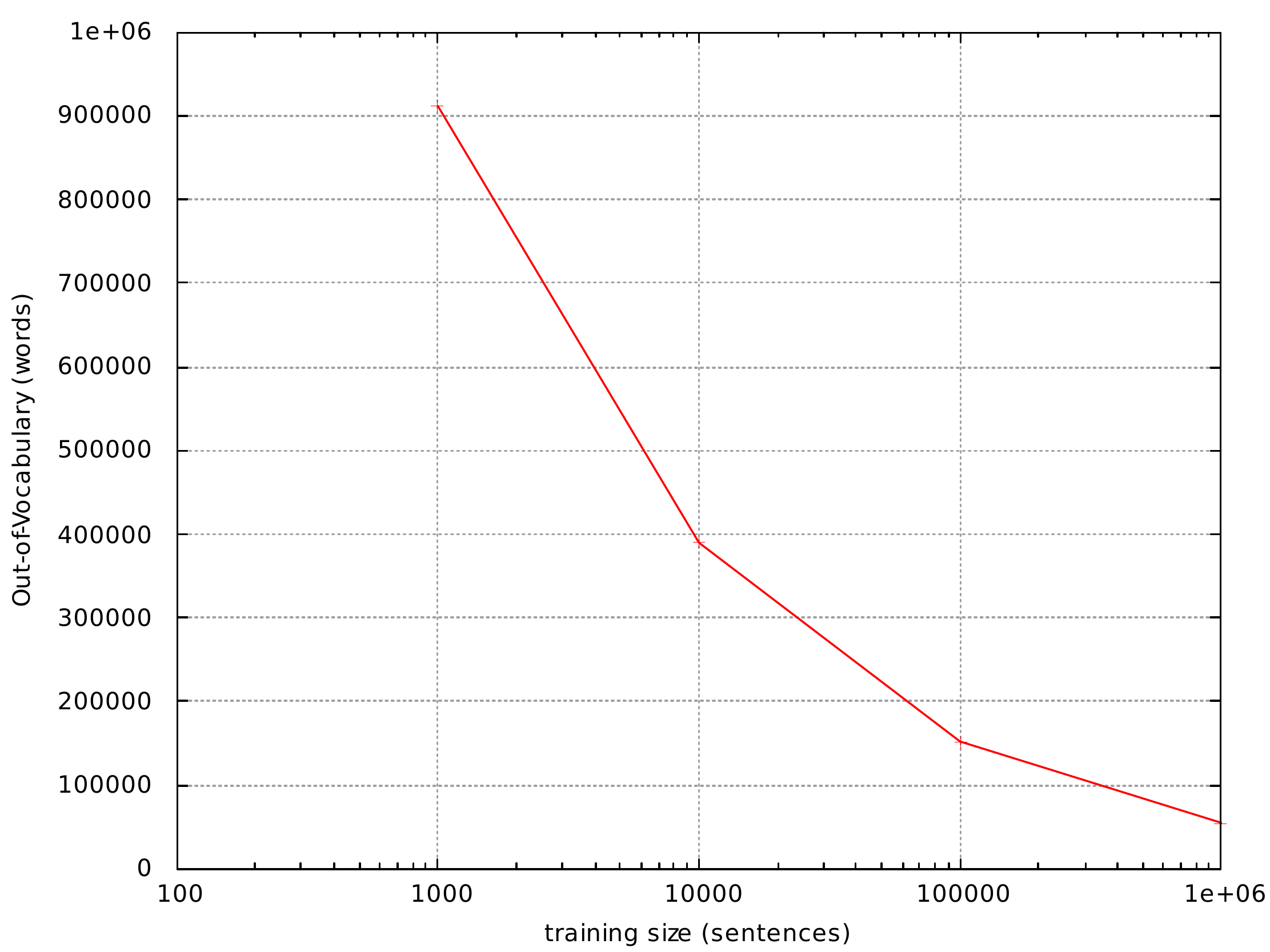}
 \caption{Number of Out-of-Vocabulary for each training set}
 \label{fig:oov}
\end{figure}

\begin{enumerate}[parsep=0pt,itemsep=0pt]
    \item We splitted text into two sets,
        i.e. training set and test set.
        We made 90:10 proportion of training and test set.
    \item We split the training sets into 4 disjoint texts i.e. texts
        with size of 1K, 10K, 100K, and 1M sentences. Total
        number of words in the training sets is about 22M.
        See Table \ref{tab:training_sets}. 
\end{enumerate}
\begin{table}
    \centering
    \caption{Number of Words and Sentences in Training sets (Tr) and Test Set(TS)}
    \begin{tabular}{| l | c | c | c | c | c |} \hline 
       Sets & TR1 & TR2 & TR3 & TR4 & TS\\ \hline \hline
       Sentences & 1K & 10K & 100K & 1M & 180K\\ \hline 
       Words & 20,436 & 204,047 & 2,057,854 & 20,567,899 & 3,696,382 \\ \hline
    \end{tabular}
    \label{tab:training_sets}
\end{table}
For each training set, we build $n$-gram language model, with $n$ = 3,5, and 7.
We then compute perplexity of one same test set for all language models.
We used following SRILM commands to build language models (example for
1M training set 7-gram language model).
\begin{enumerate}[parsep=0pt,itemsep=0pt]
    \item To build language model with Modified Kneser-Ney:
        \texttt{ngram-count -text 1M -order 7 -lm 1M.7.kn.lm -kndiscount}
    \item To build language model with Witten-Bell:
        \texttt{ngram-count -text 1M -order 7 -lm 1M.7.wb.lm -wbdiscount}
    \item To count perplexity (Modified Kneser-Ney):
        \texttt{ngram -ppl test -order 7 -lm 1M.7.kn.lm}
\end{enumerate}
Number of OOV for each training set is shown in
Fig. \ref{fig:oov}.

It is important to note here that We did not attempt
to optimise parameters in 
language models. 
In our experiment, We used default parameters values for both techniques.

\begin{figure}
 \centering
 \includegraphics[width=10cm]{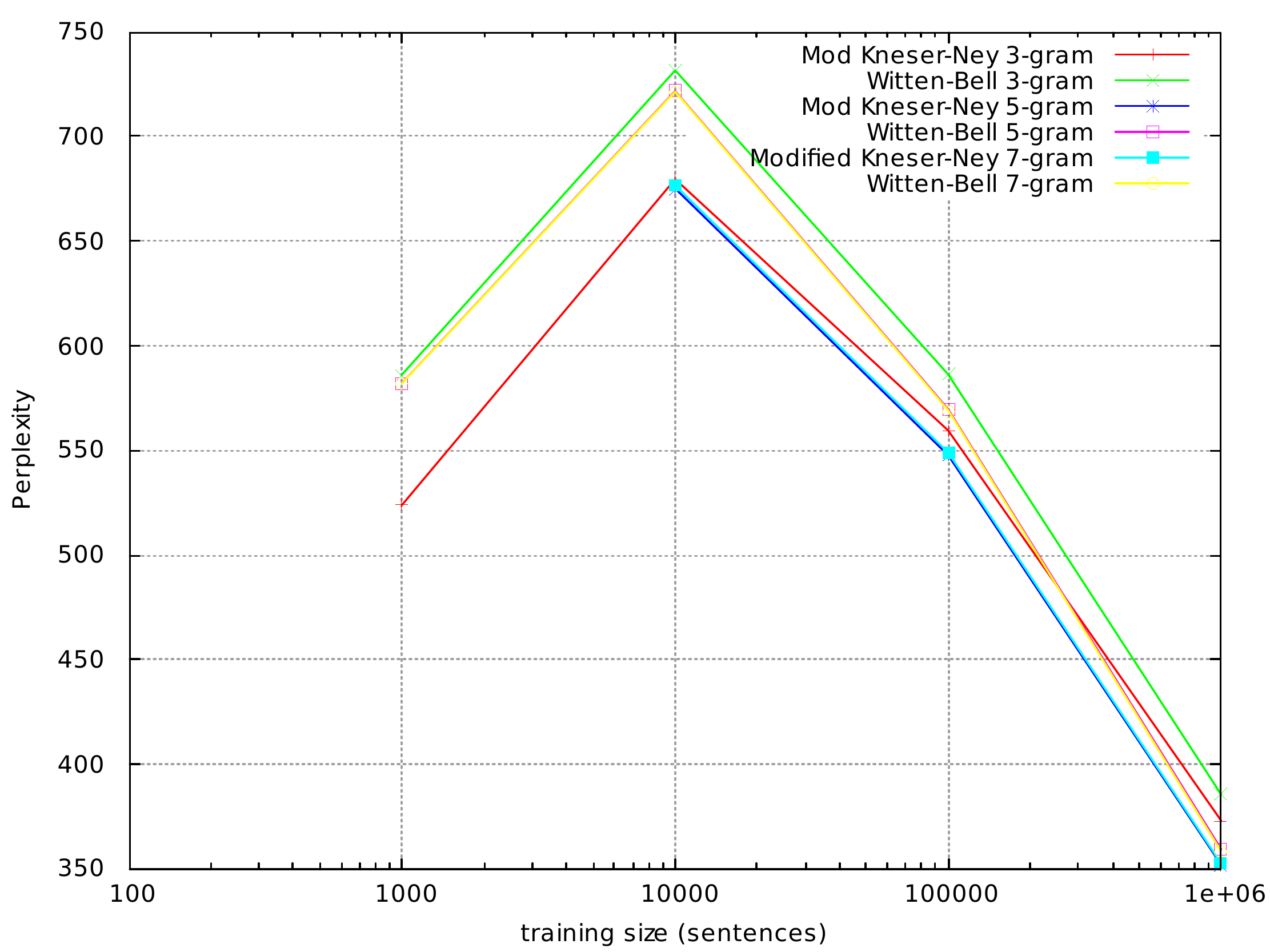}
 \caption{Perplexity of Modified Kneser-Ney and Witten-Bell for various $n$-grams and training sizes}
 \label{fig:perplexity}
\end{figure}

\begin{figure}
 \centering
 \includegraphics[width=10cm]{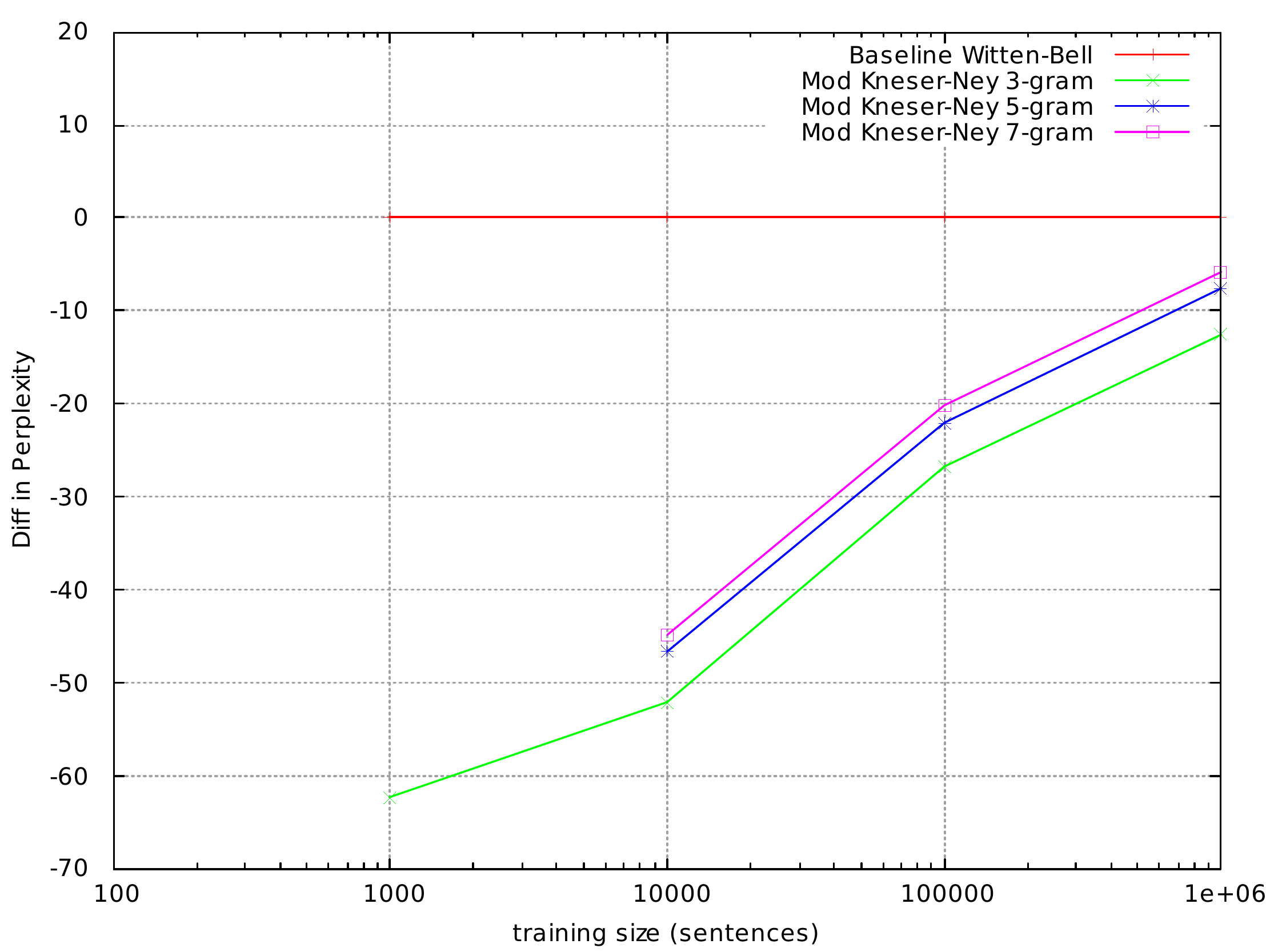}
 \caption{Differences in perplexities. Witten-Bell as a baseline. Each n-gram Modified
 Kneser-Ney is compared to its corresponding n-gram Witten-Bell.}
 \label{fig:diff}
\end{figure}

The results of our experiments are shown in Fig. \ref{fig:perplexity}.
We also plotted differences between perplexity of two smoothing techniques
with Witten-Bell as a baseline in Fig. \ref{fig:diff}.

It is interesting to note that Modified Kneser-Ney 5-gram outperforms
Modified Kneser-Ney 7-gram. 
Meanwhile, Witten-Bell technique is consistently
improved over the increase of $n$-gram order.

\section{Conclusion}
We have studied Modified Kneser-Ney and Witten-Bell smoothing
techniques for language model of Bahasa Indonesia. We used
Indonesian version of Wikipedia as a source of training and test sets.
After some data preparations, we used SRILM toolkit to build language
models and calculated perplexity for each model. Our experiments
with 3-gram, 5-gram, and 7-gram models
showed that Modified Kneser-Ney consistently outperforms
Witten-Bell technique. We will study another smoothing techniques
as well as make a standard corpus in Bahasa Indonesia
for our future research.

\bibliographystyle{plain}
\bibliography{references}

\end{document}